\documentclass[conference]{IEEEtran}
\IEEEoverridecommandlockouts

\usepackage{amsmath,amssymb}
\usepackage{graphicx}
\usepackage[hidelinks]{hyperref}
\usepackage{tikz}
\usepackage{pgfplots}
\usepackage{booktabs}
\usepackage{tabularx}
\usepackage{multirow}
\usepackage{caption}
\usepackage{subcaption}
\usepackage{orcidlink}

\pgfplotsset{compat=1.18}
\usetikzlibrary{positioning,arrows.meta,calc,fit,backgrounds,shapes.geometric,shapes.misc}

\begin{document}

\title{Native Explainability for Bayesian Confidence Propagation Neural Networks: A Framework for Trusted Brain-Like AI}

\author{
\IEEEauthorblockN{Georgios Makridis\,\orcidlink{0000-0002-6165-7239}\IEEEauthorrefmark{1}\IEEEauthorrefmark{2},
Georgios Fatouros\,\orcidlink{0000-0001-6843-089X}\IEEEauthorrefmark{1},
John Soldatos\,\orcidlink{0000-0002-6668-3911}\IEEEauthorrefmark{3},
George Katsis\,\IEEEauthorrefmark{1},
Dimosthenis Kyriazis\,\orcidlink{0000-0001-7019-7214}\IEEEauthorrefmark{2}}
\IEEEauthorblockA{
\IEEEauthorrefmark{1}ExpertAI-Lux S.à r.l,
\IEEEauthorrefmark{2}University of Piraeus, Greece, Department of Digital Systems\\
\IEEEauthorrefmark{3}University of Glasgow School of Computing Science\\
\{george.makridis, george.fatouros, george.katsis\}@expertai-lux.com, \{gmakridis, dimos\}@unipi.gr, John.Soldatos@glasgow.ac.uk}
}
\maketitle

\begin{abstract}
The EU Artificial Intelligence Act (Regulation 2024/1689), fully applicable to high-risk systems from August 2026, creates urgent demand for AI architectures that are simultaneously trustworthy, transparent, and feasible to deploy on resource-constrained edge devices. Brain-like neural networks built on the Bayesian Confidence Propagation Neural Network (BCPNN) formalism have re-emerged as a credible alternative to backpropagation-driven deep learning. They deliver state-of-the-art unsupervised representation learning, neuromorphic-friendly sparsity, and existing FPGA implementations that target edge deployment. Despite this momentum, no systematic framework exists for explaining BCPNN decisions --- a gap the present paper fills. We argue that BCPNN is, in the sense of Rudin's interpretable-by-design agenda, an inherently transparent model whose architectural primitives map directly onto established explainable-AI (XAI) families. We make four contributions. First, we propose the first XAI taxonomy for BCPNN. It maps weights, biases, hypercolumn posteriors, structural-plasticity usage scores, attractor dynamics, and input-reconstruction populations onto attribution, prototype, concept, counterfactual, and mechanistic explanation modalities. Second, we introduce sixteen architecture-level explanation primitives (P1--P16), several without analogue in standard ANNs. We provide closed-form algorithms for computing each from quantities the model already maintains. Third, we introduce five design-time Configuration-as-Explanation primitives (Config-P1 to Config-P5) that treat BCPNN hyperparameter choices as an auditable pre-deployment explanation artifact. Fourth, we sketch a roadmap for integration into industrial IoT deployments and discuss EU AI Act alignment, edge feasibility, and Industry 5.0 implications.
\end{abstract}


\section{Introduction}\label{sec:intro}

The convergence of the Internet-of-Things (IoT), AI, and cyber-physical systems is reshaping industrial decision-making. The European Industry~5.0 paradigm promotes a human-centric, sustainable and resilient technological transition in which AI systems augment rather than replace human expertise~\cite{rozanec2022humancentric}. These systems must operate at the edge under strict energy budgets, while satisfying mounting regulatory expectations on transparency. The EU AI Act (Regulation (EU) 2024/1689) entered into force in August 2024. Article~13 mandates that high-risk AI systems be ``designed and developed in such a way as to ensure that their operation is sufficiently transparent to enable deployers to interpret a system's output and use it appropriately''~\cite{euaiact2024}. The Act becomes fully applicable to high-risk systems in August 2026.

This regulatory horizon collides with two technical trends. On one hand, deep learning models have expanded into safety-critical domains while remaining opaque to all but the most sophisticated post-hoc analysis~\cite{arrieta2020xai,saeed2023xaisurvey}. On the other hand, neuromorphic and brain-like computing have matured into viable alternatives that natively implement sparsity, locality, and probabilistic inference~\cite{ravichandran2025neurocomputing,ravichandran2024spiking,hafiz2025arc}. Among these, the Bayesian Confidence Propagation Neural Network (BCPNN)~\cite{lansner1989bcpnn,sandberg2002attractor,lansner2009associative} stands out. It is grounded in Bayesian inference, learns through local Hebbian rules, produces sparse distributed representations via hypercolumn-organised soft winner-take-all (WTA) competition, and runs on commercial FPGAs with sub-watt energy budgets~\cite{hafiz2025arc}.

Despite this momentum, no systematic framework for explaining BCPNN decisions exists --- a gap the present paper fills.

\textbf{Contributions.} We argue that BCPNN is, in the sense of Rudin's interpretable-by-design agenda~\cite{rudin2019stop}, an inherently transparent architecture. Each weight is a point-wise mutual information, each bias is a log-prior, each hypercolumn output is a calibrated posterior over a discrete attribute, and the sparse patchy connectivity emerges from a mutual-information-driven structural plasticity rule. The dominant XAI families (attribution, prototype, concept, counterfactual, mechanistic) are therefore supported either out of the box or with very small adaptations.

\begin{itemize}
\item \textbf{Native XAI taxonomy.} We propose the first systematic XAI taxonomy for BCPNN, organised along the (scope, stage, model-dependency) axes established by recent surveys~\cite{arrieta2020xai,guidotti2018survey,speith2022taxonomies}.

\item \textbf{Sixteen architecture-level primitives.} We identify sixteen BCPNN-specific explanation primitives (P1--P16), several of which have no direct analogue in standard ANNs. We provide closed-form algorithms for computing each from quantities the model already maintains.

\item \textbf{Configuration-as-Explanation.} We introduce five design-time primitives (Config-P1 to Config-P5) that treat the BCPNN's hyperparameter choices as an auditable pre-deployment explanation artifact, enabling EU AI Act Article~13 compliance documentation to be generated before training begins (Section~\ref{sec:config}).

\item \textbf{Roadmap.} We sketch a roadmap for integration into industrial IoT deployments, with EU AI Act alignment and edge feasibility analyses (Section~\ref{sec:discussion}).
\end{itemize}

The paper is organised as follows. Section~\ref{sec:background} reviews XAI taxonomies and the BCPNN formalism. Section~\ref{sec:taxonomy} presents the proposed mapping. Section~\ref{sec:algorithms} gives algorithms for computing each architecture-level primitive (P1--P16). Section~\ref{sec:config} introduces the Configuration-as-Explanation primitives. Section~\ref{sec:discussion} discusses regulatory and edge implications. Section~\ref{sec:challenges} states open challenges. Section~\ref{sec:conclusion} concludes.

\section{Background and Related Work}\label{sec:background}

\subsection{Explainable AI: families and taxonomy}

Modern XAI surveys agree on three orthogonal axes for organising explanation methods: scope (local vs.\ global), stage (intrinsic vs.\ post-hoc), and model dependency (model-specific vs.\ model-agnostic)~\cite{arrieta2020xai,saeed2023xaisurvey,guidotti2018survey,speith2022taxonomies}. Within these axes, five method families dominate the literature.

\textbf{Attribution methods} assign an importance score to each input feature for a prediction; representative approaches are SHAP~\cite{lundberg2017shap}, LIME~\cite{ribeiro2016lime}, Integrated Gradients~\cite{sundararajan2017ig}, layer-wise relevance propagation~\cite{bach2015lrp}, and DeepLIFT~\cite{shrikumar2017deeplift}. Attribution methods adapted to time-series have also been developed, combining LIME and Grad-CAM for multivariate signals~\cite{makridis2024timeseries}.

\textbf{Prototype-based methods} explain decisions through similarity to learned exemplars; ProtoPNet~\cite{chen2019protopnet} introduced the ``this looks like that'' paradigm.

\textbf{Concept-based methods} ground predictions in human-interpretable variables. TCAV~\cite{kim2018tcav} quantifies sensitivity along user-specified concept directions; Concept Bottleneck Models~\cite{koh2020cbm} structurally enforce a concept layer between input and output and have spawned substantial follow-up work on concept leakage~\cite{galliamov2025cibm}.

\textbf{Counterfactual explanations} answer ``what input change would have changed the decision?''; Wachter et al.~\cite{wachter2018counterfactual} grounded the approach in the GDPR's Article~22 right-to-explanation, and the EU AI Act's Article~13 has further accentuated their relevance for high-risk decisions.

\textbf{Mechanistic interpretability} decomposes dense activations into sparse, monosemantic latent codes. Sparse autoencoders~\cite{bricken2023monosemanticity} have become the dominant tool for retrieving interpretable features from large transformers, but they are computationally heavy and post-hoc.

A recurring observation in this literature is that intrinsic, model-specific interpretability is preferable to post-hoc surrogates for high-stakes decisions~\cite{rudin2019stop,koh2020cbm}. We place BCPNN firmly in this corner.

\subsection{The Bayesian Confidence Propagation Neural Network}

BCPNN originated in Lansner and Ekeberg's derivation of a Bayesian-Hebbian learning rule~\cite{lansner1989bcpnn} and has been progressively extended to support attractor dynamics~\cite{sandberg2002attractor}, biophysical detail~\cite{lansner2009associative}, and modern unsupervised representation learning at competitive accuracy on standard benchmarks~\cite{ravichandran2025neurocomputing,ravichandran2024spiking}. Recent neuromorphic implementations on SpiNNaker~\cite{knight2016spinnaker}, FPGAs~\cite{podobas2021streambrain,hafiz2025arc}, and embedded platforms have brought BCPNN into the regime where edge deployment is not merely theoretical, achieving sub-watt inference on embedded FPGA platforms~\cite{hafiz2025arc}.

\textbf{Architecture.} A BCPNN layer is organised into $H$ hypercolumns, each containing $M$ minicolumns. Within each hypercolumn, activities are normalised by soft-WTA competition,
\begin{equation}\label{eq:wta}
\pi_{jk} = \frac{\exp(s_{jk})}{\sum_{\ell=1}^{M_{\mathrm{hid}}} \exp(s_{j\ell})},
\end{equation}
yielding a discrete probability distribution over the $M_j$ minicolumns of hypercolumn~$j$. The total support received by minicolumn $(j,k)$ decomposes additively across input hypercolumns (see Appendix~\ref{app:eqs}).

\textbf{Bayesian-Hebbian learning.} Synaptic and bias updates derive from probability traces (the p-traces) of marginal and joint activations:
\begin{align}
b_{jk} &= \log p_{jk}, \label{eq:bias}\\
w_{imjk} &= \log \frac{p_{imjk}}{p_{im}\,p_{jk}}. \label{eq:weight}
\end{align}
Equation~\eqref{eq:bias} expresses the bias as the self-information (surprisal) of the post-synaptic minicolumn~\cite{ravichandran2025neurocomputing,lansner1989bcpnn,sandberg2002attractor}. In other words, each weight records whether the pre- and post-synaptic minicolumns co-activate above chance (positive), are statistically independent (zero), or avoid each other (negative). This is precisely the quantity that SHAP and LRP attempt to estimate post-hoc through expensive surrogate procedures; BCPNN computes it exactly as a side-effect of training.

\textbf{Structural plasticity.} The connection mask $c_{ij} \in \{0,1\}$ is updated by an activity-dependent rule that maximises a usage score
\begin{equation}\label{eq:usage}
U_{ij} = \frac{\sum_{m,k} p_{imjk}\,w_{imjk}}{\sum_{k} c_{ik}}.
\end{equation}
The usage score $U_{ij}$ is the normalised mutual information between hypercolumn pair $(i,j)$; sorting it produces a ranked global feature-importance graph without any perturbation experiments. Active and silent connections are swapped when their usages differ by a factor exceeding a threshold $\rho$, yielding sparse patchy connectivity.

\textbf{Hybrid architecture.} For pattern completion, perceptual rivalry, and distortion resistance, the hybrid BCPNN adds within-layer recurrent connections in the hidden population (HID) and a feedback population (INPRC) reconstructing the input from the hidden representation~\cite{ravichandran2024spiking}. The recurrent dynamics implement attractor-based associative memory.

\textbf{Spiking variant.} A direct spiking translation~\cite{ravichandran2024spiking} replaces the rate-based activations with Poisson samples and introduces z-traces (short-term filtering of pre- and post-synaptic spikes) and p-traces (long-term probability estimates) that recover the rate-based learning rule in expectation, enabling deployment on neuromorphic hardware.

\subsection{Interpretability for related architectures}

To the best of our knowledge, no published work proposes a systematic XAI framework for BCPNN. The closest neighbours are: feature-attribution methods adapted to spiking neural networks~\cite{bitar2023snn,nguyen2023temporalspike,kim2021spikeintervals}; concept bottleneck models~\cite{koh2020cbm}, whose architectural pattern matches BCPNN's hypercolumn structure; and sparse-autoencoder-based mechanistic interpretability~\cite{bricken2023monosemanticity}, whose monosemantic features are conceptually mirrored by BCPNN's minicolumns. Recent work on user-centric XAI evaluation~\cite{makridis2025virtualxai} further motivates a per-attribute factorised explanation. We take inspiration from all three lines while exploiting the fact that BCPNN's probabilistic semantics make most adaptations unnecessary.

\section{An Explainability Taxonomy for BCPNN}\label{sec:taxonomy}

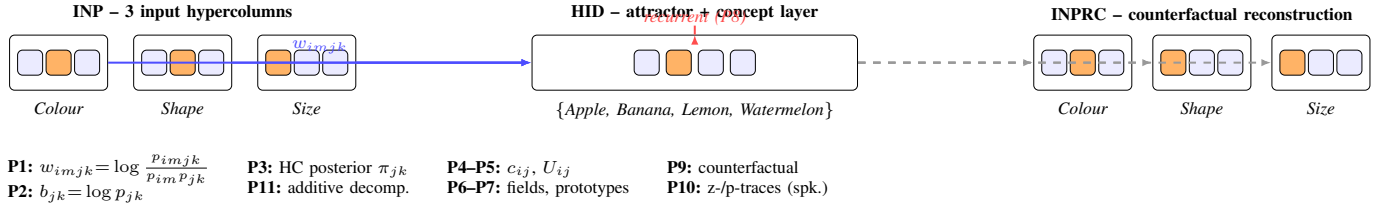
\begin{figure*}[!t]
\centering
\resizebox{\textwidth}{!}{%
\begin{tikzpicture}[
  node distance=4pt,
  every node/.style={font=\footnotesize},
  hcbox/.style={draw, rounded corners=2pt, inner sep=3pt, minimum height=22pt},
  micol/.style={draw, rectangle, minimum width=10pt, minimum height=10pt, inner sep=0pt},
  win/.style={micol, fill=orange!60},
  lose/.style={micol, fill=blue!8},
  ffarrow/.style={-{Latex[length=4pt]}, blue!70, thick},
  rcarrow/.style={-{Latex[length=4pt]}, red!70, thick},
  fbarrow/.style={-{Latex[length=4pt]}, dashed, gray!80, thick},
  lbl/.style={font=\scriptsize\itshape}
]
\node[hcbox] (col1) {\begin{tikzpicture}[baseline=0pt]
  \node[lose] (a1) at (0,0) {}; \node[win] (a2) at (0.40,0) {}; \node[lose] (a3) at (0.80,0) {};
\end{tikzpicture}};
\node[lbl, below=1pt of col1] {Colour};

\node[hcbox, right=10pt of col1] (col2) {\begin{tikzpicture}[baseline=0pt]
  \node[lose] (b1) at (0,0) {}; \node[win] (b2) at (0.40,0) {}; \node[lose] (b3) at (0.80,0) {};
\end{tikzpicture}};
\node[lbl, below=1pt of col2] {Shape};

\node[hcbox, right=10pt of col2] (col3) {\begin{tikzpicture}[baseline=0pt]
  \node[win] (c1) at (0,0) {}; \node[lose] (c2) at (0.40,0) {}; \node[lose] (c3) at (0.80,0) {};
\end{tikzpicture}};
\node[lbl, below=1pt of col3] {Size};

\node[draw=none, above=2pt of col2, font=\scriptsize\bfseries] (inplab) {INP -- 3 input hypercolumns};

\node[hcbox, right=70pt of col3, minimum width=130pt] (hid)
  {\begin{tikzpicture}[baseline=0pt]
    \node[lose] (h1) at (0,0) {}; \node[win] (h2) at (0.45,0) {}; \node[lose] (h3) at (0.90,0) {}; \node[lose] (h4) at (1.35,0) {};
   \end{tikzpicture}};
\node[lbl, below=1pt of hid] {\{Apple, Banana, Lemon, Watermelon\}};
\node[draw=none, above=2pt of hid, font=\scriptsize\bfseries] {HID -- attractor + concept layer};

\path (hid.north) edge[rcarrow, out=80, in=100, looseness=4] node[lbl, above]{\scriptsize recurrent (P8)} (hid.north);

\node[hcbox, right=70pt of hid] (rc1) {\begin{tikzpicture}[baseline=0pt]
  \node[lose] (rc1a) at (0,0) {}; \node[win] (rc1b) at (0.40,0) {}; \node[lose] (rc1c) at (0.80,0) {};
\end{tikzpicture}};
\node[lbl, below=1pt of rc1] {Colour};

\node[hcbox, right=8pt of rc1] (rc2) {\begin{tikzpicture}[baseline=0pt]
  \node[win] (rc2a) at (0,0) {}; \node[lose] (rc2b) at (0.40,0) {}; \node[lose] (rc2c) at (0.80,0) {};
\end{tikzpicture}};
\node[lbl, below=1pt of rc2] {Shape};

\node[hcbox, right=8pt of rc2] (rc3) {\begin{tikzpicture}[baseline=0pt]
  \node[win] (rc3a) at (0,0) {}; \node[lose] (rc3b) at (0.40,0) {}; \node[lose] (rc3c) at (0.80,0) {};
\end{tikzpicture}};
\node[lbl, below=1pt of rc3] {Size};
\node[draw=none, above=2pt of rc2, font=\scriptsize\bfseries] {INPRC -- counterfactual reconstruction};

\draw[ffarrow] (col1.east) -- node[lbl, above, sloped] {$w_{imjk}$} (hid.west);
\draw[ffarrow] (col2.east) -- (hid.west);
\draw[ffarrow] (col3.east) -- (hid.west);

\draw[fbarrow] (hid.east) -- (rc1.west);
\draw[fbarrow] (hid.east) -- (rc2.west);
\draw[fbarrow] (hid.east) -- (rc3.west);

\node[draw=none, font=\scriptsize, align=left, below=22pt of col1.south west, anchor=north west, xshift=-4pt] (a1lbl) {\textbf{P1:} $w_{imjk}{=}\log\frac{p_{imjk}}{p_{im}p_{jk}}$\\\textbf{P2:} $b_{jk}{=}\log p_{jk}$};
\node[draw=none, font=\scriptsize, align=left, right=8pt of a1lbl] (a2lbl) {\textbf{P3:} HC posterior $\pi_{jk}$\\\textbf{P11:} additive decomp.};
\node[draw=none, font=\scriptsize, align=left, right=8pt of a2lbl] (a3lbl) {\textbf{P4--P5:} $c_{ij}$, $U_{ij}$\\\textbf{P6--P7:} fields, prototypes};
\node[draw=none, font=\scriptsize, align=left, right=8pt of a3lbl] (a4lbl) {\textbf{P9:} counterfactual\\\textbf{P10:} z-/p-traces (spk.)};
\end{tikzpicture}}
\caption{Schematic of a BCPNN annotated with the architecture-level explanation primitives proposed in this paper. Orange minicolumns denote winners under soft-WTA competition (Eq.~\ref{eq:wta}). Solid blue arrows are feedforward, the curved red arrow is recurrent (HID$\leftrightarrow$HID), and dashed arrows are feedback to the input-reconstruction (INPRC) population. Each primitive is computed in closed form from quantities the network already maintains during inference --- no surrogate models, gradients or perturbations are required.}
\label{fig:overview}
\end{figure*}

We now state the central thesis of the paper: each architectural element of the BCPNN doubles as a native explanation primitive. Table~\ref{tab:taxonomy} summarises the mapping; Fig.~\ref{fig:overview} provides a schematic.

\subsection{Why BCPNN weights are explanations, not parameters}

In a standard ANN, a weight is an opaque scalar whose interpretation is contingent on the surrounding network state. In BCPNN, equation~\eqref{eq:weight} guarantees that each weight is, by construction, a point-wise mutual information --- positive for above-chance co-occurrence, zero for independence, negative for dis-association~\cite{ravichandran2025neurocomputing}. Since point-wise mutual information is precisely the quantity SHAP~\cite{lundberg2017shap}, LRP~\cite{bach2015lrp} and TCAV~\cite{kim2018tcav} attempt to estimate post-hoc, BCPNN computes the explanation as a side-effect of training, and combined with biases as log-priors~\eqref{eq:bias} the support computation reads as a Bayesian log-likelihood update~\cite{rudin2019stop,wachter2018counterfactual}.

\subsection{Why hypercolumn outputs solve the calibration problem}

A persistent issue in standard ANN explainability is that softmax outputs are generally not probabilities; modern deep networks are systematically over-confident. In BCPNN, equation~\eqref{eq:wta} produces a calibrated discrete posterior by construction, since the soft-WTA dynamics implement a normalised likelihood ratio test under the BCPNN generative model~\cite{lansner1989bcpnn,sandberg2002attractor}. Each hypercolumn therefore reports its own per-attribute confidence, satisfying the AI Act Article~13 requirement that systems convey ``the level of accuracy''~\cite{euaiact2024} on a per-attribute basis.

\subsection{Why structural plasticity is intrinsic feature selection}

The connectivity mask $c_{ij}$ and the usage score $U_{ij}$ jointly perform feature selection at the hypercolumn-pair level (P4, P5). Unlike post-hoc methods that estimate global feature importance through perturbation or gradient analysis~\cite{arrieta2020xai,guidotti2018survey}, BCPNN's feature selection is an intrinsic byproduct of the structural plasticity rule. The resulting bipartite graph between input and hidden hypercolumns is the global explanation of which input attributes the network deems relevant.

\subsection{Why minicolumns are prototypes}

Within each hypercolumn, soft-WTA competition~\eqref{eq:wta} ensures that typically a single minicolumn dominates per input. The activation profile of each minicolumn over the data distribution --- a tuning curve over input states --- behaves as a learned prototype, achieving the ``this looks like that'' semantics of ProtoPNet~\cite{chen2019protopnet} without any auxiliary prototype loss. The concept-leakage problem that plagues Concept Bottleneck Models~\cite{koh2020cbm,galliamov2025cibm} is suppressed by the soft-WTA quantisation: information that does not project onto the leading minicolumn is discarded.

\subsection{Why attractor dynamics expose decision paths}

In the hybrid BCPNN, the recurrent dynamics let the network state evolve from a feedforward-driven initial estimate to a clean attractor, and the trajectory through state-space (P8) is itself an explanation --- a long oscillatory path indicates perceptual rivalry, a short direct path high confidence. The INPRC feedback population (P9) provides an in-distribution generative counterfactual --- ``the network thinks the input should look like this'' --- satisfying the counterfactual explanation requirement of Wachter et al.~\cite{wachter2018counterfactual} without external generative models.

\begin{table*}[!t]
\centering
\caption{Mapping between BCPNN structural elements and XAI families. Asterisks ($\ast$) mark primitives with no direct analogue in standard ANNs.}
\label{tab:taxonomy}
\renewcommand{\arraystretch}{1.15}
\begin{tabularx}{\textwidth}{@{}l X l l@{}}
\toprule
\textbf{BCPNN element} & \textbf{Native explanation it provides} & \textbf{XAI family} & \textbf{Closed-form quantity}\\
\midrule
\multicolumn{4}{l}{\emph{Architecture-level primitives (P1--P11)}}\\
\midrule
P1.\ Weight $w_{imjk}$ & Pointwise evidence of $(j,k)$ given $(i,m)$ & Attribution, mech.\ interp. & $\log p_{imjk}/(p_{im}p_{jk})$\\
P2.\ Bias $b_{jk}$ & Surprisal / log-prior of minicolumn $(j,k)$ & Attribution baseline & $\log p_{jk}$\\
P3.\ Hypercolumn posterior & Calibrated per-attribute uncertainty & Uncertainty-aware & $\pi_{jk}$ via Eq.~\eqref{eq:wta}\\
P4.\ Connectivity $c_{ij}$ & Which input attribute drives which concept & Concept-level FS$^{\ast}$ & Binary support graph\\
P5.\ Usage score $U_{ij}$ & Normalised MI per HC--HC connection & Global concept importance$^{\ast}$ & Eq.~\eqref{eq:usage}\\
P6.\ Receptive fields & Native per-minicolumn saliency & Saliency / TCAV-like & Aggregated $w_{imjk}$\\
P7.\ Minicolumn tunings & Built-in prototypes per attribute & Prototype-based & Activation tuning curves\\
P8.\ Attractor trajectory & Path-of-decision through state-space$^{\ast}$ & Dynamical-systems & $\{\pi_{jk}(t)\}_{t \le T}$\\
P9.\ INPRC reconstruction & In-distribution counterfactual & Counterfactual & Generative INPRC output\\
P10.\ z-/p-traces (spk.) & Per-time-step evidence accumulation & Temporal saliency$^{\ast}$ & Eqs.\ of \cite{ravichandran2024spiking}\\
P11.\ Modular factorisation & Per-hypercolumn additive explanation & Additive / GAM-like & $\sum_h \log \pi^{(h)}_{jk}$\\
\midrule
\multicolumn{4}{l}{\emph{New architecture-level primitives (P12--P16)}}\\
\midrule
P12.\ Surprise score & OOD detection from posterior collapse$^{\ast}$ & OOD / uncertainty & $\mathcal{S}(x)=-\sum_j \log \pi_{jk^*}(x)$\\
P13.\ P-trace drift & Concept-drift self-monitoring$^{\ast}$ & Drift detection & CUSUM over $p_{jk}$\\
P14.\ Certified WTA radius & Local robustness without gradients$^{\ast}$ & Robustness certificate & Per-HC $\delta_j$\\
P15.\ Winner margin & Per-attribute decision confidence & Local uncertainty & $\Delta_j=\pi_{jk^*}-\pi_{jk^{(2)}}$\\
P16.\ Cross-layer attribution & Exact deep-BCPNN attribution path$^{\ast}$ & Deep attribution & Chained $\phi_{i\to jk}$\\
\midrule
\multicolumn{4}{l}{\emph{Design-time (Configuration) primitives (Config-P1--Config-P5)}}\\
\midrule
Config-P1.\ Ontology declaration$^{\ast}$ & Pre-training auditable scope $(H,\{M_j\},\{c_{ij}\})$ & System-level pre-hoc & Static document\\
Config-P2.\ Configuration efficiency$^{\ast}$ & Detection of capacity mismatch & Architecture audit & $\overline{\mathrm{Diff}}$ over HCs\\
Config-P3.\ Plasticity threshold $\rho$ & Interpretability/accuracy tradeoff knob & Native Pareto control & $\rho$ sweep curve\\
Config-P4.\ Configuration fidelity$^{\ast}$ & Domain-prior validation & Ontology validation & $\mathrm{Spearman}$ test\\
Config-P5.\ Temporal scope $\tau_z$ & Auditable temporal-memory window & Temporal scoping & $\tau_z$ specification\\
\bottomrule
\end{tabularx}
\end{table*}

\section{Computing Explanations from BCPNN Quantities}\label{sec:algorithms}

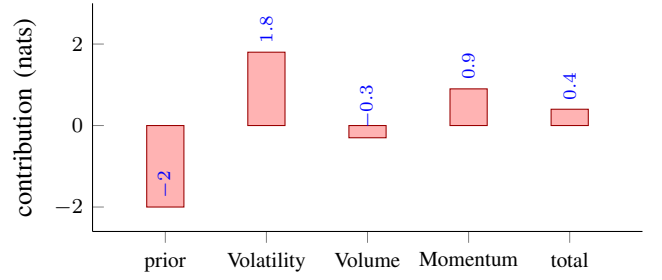
\begin{figure}[!t]
\centering
\begin{tikzpicture}
\begin{axis}[
  ybar,
  width=\columnwidth,
  height=4.6cm,
  ylabel={contribution (nats)},
  symbolic x coords={prior, Volatility, Volume, Momentum, total},
  xtick=data,
  xticklabel style={font=\footnotesize},
  yticklabel style={font=\footnotesize},
  ymin=-2.6, ymax=3.0,
  bar width=14pt,
  enlarge x limits=0.18,
  nodes near coords,
  nodes near coords style={font=\scriptsize},
  every node near coord/.append style={anchor=west, rotate=90},
  axis y line*=left,
  axis x line*=bottom,
]
\addplot+[fill=red!30, draw=red!60!black] coordinates {
  (prior,-2.0) (Volatility,1.8) (Volume,-0.3) (Momentum,0.9) (total,0.4)
};
\end{axis}
\end{tikzpicture}
\caption{P11 (Modular Factorisation): illustrative per-hypercolumn support decomposition for a financial risk query. Each bar is the additive contribution of one input attribute to the winning hidden minicolumn, computed in closed form from Eq.~\eqref{eq:bias} and Eq.~\eqref{eq:weight} without any surrogate model. Values are illustrative.}
\label{fig:additive}
\end{figure}

We now state algorithms for computing each primitive. All quantities are already maintained by a trained BCPNN; no surrogate models, gradient back-propagation, or perturbation budgets are required. We denote a query input by $x$, with $\pi_{im}(x)$ the activity of minicolumn $(i,m)$ in the input layer. Figure~\ref{fig:additive} illustrates the per-hypercolumn additive support decomposition (P11) on a single-query case study.

\subsection{Local attribution (P1, P2, P11)}

Each contribution term $\phi_{i\to jk}(x)$ is the additive evidence supplied by input hypercolumn $i$ to the support for hidden minicolumn $(j,k)$, expressed in nats. The per-hypercolumn additive decomposition (P11) follows directly from the independence structure of the BCPNN generative model; the closed-form expression is given in the Appendix. Each $\phi_{i\to jk}(x)$ is the quantity SHAP attempts to estimate, with two key differences. It is computed in $\mathcal{O}(H_{\mathrm{inp}}M_{\mathrm{inp}})$ rather than exponentially, and it is exact rather than approximate~\cite{lundberg2017shap}.

\subsection{Global concept importance (P4, P5)}

For each pair of hypercolumns $(i,j)$, the usage score $U_{ij}$ from Eq.~\eqref{eq:usage} is a one-line summary of how much input attribute $i$ contributes to latent concept $j$, averaged over the training distribution. Sorting $U_{ij}$ produces a ranked global feature-importance graph that requires no perturbation experiments.

\subsection{Receptive fields and prototypes (P6, P7)}

For each hidden minicolumn $(j,k)$, define its receptive field as the per-input-minicolumn weighted activation
\begin{equation}\label{eq:receptive}
R_{jk}(i,m) = \pi_{im}(x)\,w_{imjk}\,c_{ij},
\end{equation}
averaged or aggregated over a chosen reference set. Equation~\eqref{eq:receptive} can be visualised directly and serves both as a saliency map (P6) and a prototype profile (P7).

\subsection{Counterfactual via INPRC (P9)}

Given a query $x$, run the recurrent BCPNN dynamics for $T$ steps and read off the INPRC reconstruction $\hat{x}$. The pair $(x,\hat{x})$ is a contrastive explanation: the dimensions on which they differ are precisely those where the input deviates from the network's learned generative model. To produce a directed counterfactual, force the desired class minicolumn to dominate in HID and re-run the feedback pathway; the resulting $\hat{x}'$ is the closest in-distribution input that would have produced the alternative decision~\cite{ravichandran2024spiking}.

\subsection{Attractor-path explanation (P8)}

For the recurrent BCPNN, log the per-minicolumn activations $\pi_{jk}(t)$ for $t = 0,\dots,T$. Three scalar diagnostics --- settling time, basin width, and trajectory length --- compactly summarise decision confidence and ambiguity; full definitions are given in the Appendix.

\subsection{Temporal saliency for spiking BCPNN (P10)}

In the spiking variant~\cite{ravichandran2024spiking}, the z-trace $z_i(t)$ is a leaky integrator of pre-synaptic spikes with time constant $\tau_{zi}$. The instantaneous contribution of spike $s_i(t)$ to weight $w_{imjk}$ is $\Delta s_{jk}(t) = z_i(t)z_j(t)w_{imjk}$, plotted directly as a temporal saliency map. The construction is uniquely possible in BCPNN because z-traces are explicit state variables of the model. In standard RNNs, per-time-step attribution has to be approximated through gradient-based methods that scale linearly with sequence length~\cite{nguyen2023temporalspike}. Consider a vibration-sensor anomaly detector deployed on a manufacturing IoT edge node. When the model raises an alert, an operator needs to know which $50\,\mathrm{ms}$ window contained the offending spectral signature. P10 returns this window directly, with no extra computation. The same primitive supports root-cause analysis for fault diagnosis and post-incident regulatory reporting. P10 connects directly to Config-P5 below: $\tau_z$ pre-declares the temporal receptive field within which P10 is meaningful, so the spiking BCPNN's temporal explanation scope is auditable before deployment rather than estimated empirically afterwards.

\subsection{OOD detection via native surprise score (P12)}

For an out-of-distribution (OOD) input, weight terms $\phi_{i \to jk}(x)$ are small and the posterior collapses toward the prior $b_{jk}$. We define the surprise score
\begin{equation}\label{eq:surprise}
\mathcal{S}(x) = -\sum_j \log \pi_{jk^*}(x),
\end{equation}
where $k^*$ is the winning minicolumn in hypercolumn~$j$. A high $\mathcal{S}$ indicates OOD. This is a zero-cost OOD detector requiring no external anomaly module and is suitable for energy-constrained edge deployment.

\subsection{Concept drift detection via p-trace monitoring (P13)}

The p-traces $p_{jk}$ are running averages already maintained during inference. Significant drift of live $p_{jk}$ from training-time baselines means learned weights are becoming stale. A CUSUM control chart over p-trace values gives a built-in, parameter-free drift alarm: the model self-reports when its explanations can no longer be trusted.

\subsection{WTA robustness certificate (P14)}

The soft-WTA competition discretises inputs; perturbations that do not flip the winning minicolumn do not change the prediction or the explanation. We define the per-hypercolumn certified radius $\delta_j$ as the minimum input perturbation needed to flip the winner. An input where all $\delta_j$ are large is provably robust without adversarial training or gradient computation.

\subsection{Winner margin per-attribute uncertainty (P15)}

Define
\begin{equation}\label{eq:margin}
\Delta_j(x) = \pi_{jk^*}(x) - \pi_{jk^{(2)}}(x)
\end{equation}
as the gap between the top two minicolumn activations in hypercolumn~$j$. This gives per-attribute confidence rather than a single global uncertainty value. An expert reviewing a financial decision can see which attributes were decided confidently and which were borderline, with zero additional computation.

\subsection{Cross-layer exact attribution for deep BCPNN (P16)}

For multi-layer BCPNN, the additive support decomposition (P11) chains across layers: $\phi_{i \to jk}$ at layer~1 feeds into $\phi_{j \to lm}$ at layer~2, giving a closed-form exact attribution path from raw input to final decision. This is the BCPNN equivalent of Integrated Gradients, but exact and free; Integrated Gradients is approximate and requires hundreds of forward passes~\cite{sundararajan2017ig}.

\section{Configuration as Explanation: Design-Time Trustworthiness}\label{sec:config}

In a standard ANN, configuration choices --- layer width, depth, activation function --- are capacity decisions with no semantic meaning. In BCPNN this is fundamentally different: every configuration choice is a semantic commitment about the structure of the domain. Setting $H$ hypercolumns with $M_j$ minicolumns each asserts that the domain has $H$ distinct attributes, each with $M_j$ discrete states. The connectivity prior $c_{ij}$ asserts which input attributes may influence which latent concepts. Written before a single data point is seen, the configuration is already a human-readable ontology of what the model is permitted to represent --- something no post-hoc XAI method can provide. This design-pattern of treating model configuration as a persistent, auditable artifact has previously proven effective in personalised AI deployment~\cite{makridis2024fairyland}.

\subsection{Ontology Declaration (Config-P1)}
The configuration $(H, \{M_j\}, \{c_{ij}\})$ is an auditable ontology. For a financial risk application: $H=5$ hypercolumns mapped to \{Volatility, Volume, Momentum, Spread, Sentiment\} with $M_j \in \{3,4,3,4,5\}$ discrete states each. A compliance officer can read this configuration and certify the model's representational scope before seeing any weights or predictions, satisfying EU AI Act Article~13's requirement for documentation of intended purpose. This is a system-level, pre-hoc explanation, in contrast to SHAP or LIME, which are post-hoc and per-prediction.

\subsection{Configuration Efficiency Score (Config-P2)}
After training, compare the configured structure to the learned structure. For each hypercolumn $j$, the minicolumn differentiation score
\begin{equation}\label{eq:diff}
\mathrm{Diff}_j = \frac{1}{M_j(M_j-1)} \sum_{k \ne k'} \|\mathbf{w}_{\cdot jk} - \mathbf{w}_{\cdot jk'}\|_1
\end{equation}
measures whether the $M_j$ minicolumns learned genuinely distinct weight profiles. If $\mathrm{Diff}_j \approx 0$, the data did not support $M_j$ states for attribute $j$ and explanations from that hypercolumn are unreliable. The overall Configuration Efficiency $\overline{\mathrm{Diff}} = \frac{1}{H}\sum_j \mathrm{Diff}_j$ is a single pre-deployment quality metric unavailable in any standard ANN XAI framework.

\subsection{The Interpretability Dial: $\rho$ (Config-P3)}
The structural plasticity threshold $\rho$ is a native accuracy/interpretability tradeoff knob. Increasing $\rho$ prunes connections with low usage scores, simplifying the P4/P5 explanation graph at a small accuracy cost. Sweeping $\rho$ produces an Interpretability--Accuracy curve that lets practitioners choose an operating point appropriate for their regulatory context. Unlike post-hoc compression, $\rho$ shapes what the model learns, not just how it is described afterward.

\subsection{Configuration Fidelity (Config-P4)}
When a domain ontology is known before training --- financial risk factors, clinical biomarkers, IoT sensor channels --- Configuration Fidelity
\begin{equation}\label{eq:cf}
\mathrm{CF} = \mathrm{Spearman}\bigl(\mathrm{rank}_{\mathrm{expert}}(i),\, \mathrm{rank}_{U_{ij}}(i)\bigr)
\end{equation}
measures rank-order agreement between the domain expert's expected feature importance and the post-training usage scores $U_{ij}$. $\mathrm{CF} \approx 1$ validates the domain model; $\mathrm{CF} \approx 0$ or negative signals that the data carries different structure than the prior assumed, triggering a model audit before deployment, not after an incident. No post-hoc method can surface this signal pre-deployment.

\subsection{Temporal Scope Declaration (Config-P5, spiking BCPNN only)}
The z-trace time constant $\tau_z$ is an explicit, auditable statement of temporal memory: decisions integrate evidence over the last $\tau_z$ milliseconds. A BCPNN with $\tau_z = 50\,\mathrm{ms}$ is certifiably a short-memory detector suitable for impact events; the same architecture with $\tau_z = 2{,}000\,\mathrm{ms}$ is a context-integrating detector suitable for anomaly accumulation. In an RNN or temporal CNN, the effective receptive field must be estimated empirically; in BCPNN it is read directly from the model specification, satisfying the EU AI Act Article~13 requirement to communicate operational limitations to deployers.

\begin{table*}[!t]
\centering
\caption{Design-Time vs.\ Post-Hoc XAI.}
\label{tab:configvsposthoc}
\renewcommand{\arraystretch}{1.2}
\begin{tabularx}{\textwidth}{@{}l X X X@{}}
\toprule
\textbf{Property} & \textbf{Post-hoc XAI (SHAP, LIME)} & \textbf{BCPNN Arch.\ XAI (P1--P11)} & \textbf{BCPNN Config XAI (Config-P1--P5)}\\
\midrule
When available & After training & After training & Before and after training\\
Scope & Local/global & Local and global & System-level\\
Modifies the model? & No & No & Yes ($\rho$ knob)\\
Detects capacity mismatch? & No & No & Yes (Config-P2)\\
Validates domain prior? & No & No & Yes (Config-P4)\\
EU AI Act role & Post-hoc addendum & Technical report & Pre-deployment audit artifact\\
Runtime cost & High (exponential for SHAP) & Zero & Zero (static document)\\
\bottomrule
\end{tabularx}
\end{table*}

\section{Discussion: Trustworthiness, Edge AI, Industry 5.0}\label{sec:discussion}

\subsection{EU AI Act alignment}
Article~13 of the EU AI Act requires high-risk AI systems to be sufficiently transparent for deployers to ``interpret a system's output and use it appropriately'', and Article~14 requires effective human oversight~\cite{euaiact2024}. The BCPNN primitives map onto these requirements directly: per-hypercolumn calibrated confidence (P3) supports the ``level of accuracy'' clause; the structural-plasticity graph (P4, P5) supports the ``capabilities and limitations'' clause; counterfactuals (P9) and decision-path diagnostics (P8) support contestability~\cite{wachter2018counterfactual}. The Configuration-as-Explanation primitives further provide a pre-deployment audit trail (Config-P1, Config-P2, Config-P4) that no post-hoc method can produce. Compliance overhead at the edge is therefore minimal.

\subsection{Edge feasibility}
Post-hoc XAI pipelines such as SHAP, LIME and sparse autoencoders are typically infeasible at the edge because they require either repeated forward passes, surrogate model training, or substantial additional memory~\cite{lundberg2017shap,ribeiro2016lime,bricken2023monosemanticity}. Our framework adds zero runtime cost: every quantity is already computed during inference. The most recent embedded FPGA BCPNN accelerator achieves sub-watt inference on embedded platforms~\cite{hafiz2025arc}, and our framework therefore inherits the same energy profile.

\subsection{Industry 5.0 and human-centred AI}
Industry~5.0 emphasises human--AI collaboration over automation~\cite{rozanec2022humancentric}. The hypercolumn structure of BCPNN is suited to this paradigm: each hypercolumn corresponds to a human-interpretable attribute, and the per-hypercolumn explanations decompose decisions along axes that domain experts can audit or override. This decomposition mirrors expert reasoning more closely than monolithic feature-importance vectors do.

\subsection{Application vignettes}

\textbf{Financial time series.} For a BCPNN trained on financial indicators, Config-P1 directly declares an ontology --- $H=5$ hypercolumns over \{Volatility, Volume, Momentum, Spread, Sentiment\} --- before any data are seen. After training, P11 (modular factorisation) yields per-attribute log-evidence; in an illustrative scenario, the volatility hypercolumn might contribute $+1.8$ nats to the crash class while the volume hypercolumn contributes $-0.3$ nats. Such factorised explanations would be auditable and align with EBA and ESMA expectations on automated credit and market decisions, and they integrate naturally with risk-aware models such as DeepVaR~\cite{fatouros2023deepvar}, which already consume per-channel sentiment and volatility signals~\cite{makridis2024timeseries}. Conversational delivery of these explanations to non-technical stakeholders has been demonstrated in HumAIne-Chatbot deployments~\cite{makridis2025humaine}, and user-centric evaluation of the resulting workflow follows the protocol of VirtualXAI~\cite{makridis2025virtualxai}.

\textbf{IoT sensing and cybersecurity.} For multimodal sensor data, the structural-plasticity graph (P4) reveals which sensor channels the network has learned to rely on, supporting sensor pruning and fault diagnosis. The same primitives transfer to cyber-defence monitoring, where per-channel evidence decomposition has been shown to improve analyst trust and triage time~\cite{makridis2022cyber}.

\section{Open Challenges and Future Work}\label{sec:challenges}

Four challenges remain.
\textbf{(i) Scaling and abstraction.} The architecture-level primitives are linear in the number of hypercolumns, but human-interpretability of explanation graphs --- including the P4/P5 usage graph and the P8 attractor trajectory --- may degrade for deep BCPNN stacks; principled abstraction layers, concept hierarchies, and Config-P2 efficiency monitoring are needed to preserve explanatory value at scale.
\textbf{(ii) Faithfulness validation.} Although each primitive is computed from quantities the model uses for inference, formal faithfulness metrics should be measured on full-scale BCPNN systems jointly with partners running production BCPNN deployments.
\textbf{(iii) User studies.} Explanation quality is a human-factors question~\cite{miller2019explanation}. Controlled studies with domain experts in clinical and financial settings should benchmark per-hypercolumn explanations against SHAP and LIME on decision time, accuracy, and trust calibration.
\textbf{(iv) Empirical validation of new primitives.} Primitives P12--P16 are derived analytically but have not been benchmarked against standard baselines on full-scale BCPNN deployments; joint validation with partners maintaining large-scale BCPNN systems is a direct next step.

\section{Conclusion}\label{sec:conclusion}

We have argued that the Bayesian Confidence Propagation Neural Network is, by construction, an interpretable-by-design model whose architectural primitives correspond directly to the dominant XAI families. We proposed the first XAI taxonomy for BCPNN, mapping its structural elements onto attribution, prototype, concept, counterfactual, and mechanistic modalities. We introduced sixteen architecture-level explanation primitives (P1--P16), several without analogue in standard ANNs, each computed in closed form from quantities the network already maintains. We further introduced five design-time Configuration-as-Explanation primitives (Config-P1 to Config-P5) that turn the BCPNN's hyperparameter choices into a pre-deployment audit artifact, addressing EU AI Act Article~13 documentation requirements before training begins. We sketched a roadmap for integration into industrial IoT deployments, with edge feasibility and Industry~5.0 alignment analyses. Together, these contributions offer a credible path to AI systems that are simultaneously trustworthy, transparent, and feasible to deploy at the edge under the EU AI Act regime.

\section*{Acknowledgment}
This work has been supported by the EXTRA-BRAIN project, funded by the European Union's Horizon Europe programme under Grant Agreement No.~101135809.

\appendix
\section*{Detailed Equations}\label{app:eqs}

\textbf{Total support across input hypercolumns.} The total support received by minicolumn $(j,k)$ from a query $x$ decomposes additively across input hypercolumns through the BCPNN generative independence structure:
\begin{equation}\label{eq:support-app}
s_{jk}(x) = \underbrace{b_{jk}}_{\text{prior}} + \sum_{i=1}^{H_{\mathrm{inp}}}
\underbrace{\left[\sum_{m=1}^{M_{\mathrm{inp}}} \pi_{im}(x)\,w_{imjk}\,c_{ij}\right]}_{\phi_{i\to jk}(x)}.
\end{equation}

\textbf{Per-hypercolumn additive decomposition (P11).} The contribution of input hypercolumn $i$ to the support for minicolumn $(j,k)$ is the bracketed term $\phi_{i\to jk}(x)$ in Eq.~\eqref{eq:support-app}. Expressed in nats, $\phi_{i\to jk}(x)$ is the BCPNN counterpart of the SHAP attribution; it is computed in $\mathcal{O}(H_{\mathrm{inp}}M_{\mathrm{inp}})$ and is exact rather than sampled.

\textbf{Attractor-path diagnostics.} For the recurrent BCPNN, log the per-minicolumn activations $\pi_{jk}(t)$ for $t=0,\dots,T$. Three scalar diagnostics summarise the decision trajectory:
\begin{align}
T^{\star} &= \min\{t : \|\pi(t+1)-\pi(t)\|_{\infty} < \epsilon\},\label{eq:settling}\\
\beta &= -\log \pi^{\mathrm{2nd}}(T^{\star}),\label{eq:basin}\\
L &= \sum_{t=0}^{T-1} \|\pi(t+1)-\pi(t)\|_2,\label{eq:length}
\end{align}
where $\pi^{\mathrm{2nd}}$ is the second-largest minicolumn activation in the dominant hypercolumn at convergence. $T^{\star}$ is the settling time, $\beta$ is the basin width, and $L$ is the trajectory length. A larger $\beta$ corresponds to a greater gap between the winner and runner-up at convergence, indicating a deeper energy minimum and therefore stronger attractor pull --- we adopt this as a proxy for basin width in the discrete posterior space. Together they compactly summarise decision confidence and ambiguity for primitive P8.

\bibliographystyle{IEEEtran}
\bibliography{references}

\end{document}